# Automated Meta Prompt Engineering for Alignment with the Theory of Mind

Aaron Baughman
IBM
Cary NC, USA
baaron@us.ibm.com

Rahul Agarwal
IBM
New York NY, USA
rahul.agarwal@ibm.com

Eduardo Morales
IBM
Coral Gables FL, USA
Eduardo.Morales@ibm.com

Gozde Akay
IBM
Fredericton, Canada
Gozde@ibm.com

**Abstract**

We introduce a method of meta-prompting that jointly produces fluent text for complex tasks while optimizing the similarity of neural states between a human's mental expectation and a Large Language Model's (LLM) neural processing. A technique of agentic reinforcement learning is applied, in which an LLM as a Judge (LLMaaJ) teaches another LLM, through in-context learning, how to produce content by interpreting the intended and unintended generated text traits. To measure human mental beliefs around content production, users modify long form AI-generated text articles before publication at the US Open 2024 tennis Grand Slam. Now, an LLMaaJ can solve the Theory of Mind (ToM) alignment problem by anticipating and including human edits within the creation of text from an LLM. Throughout experimentation and by interpreting the results of a live production system, the expectations of human content reviewers had 100% of alignment with AI 53.8% of the time with an average iteration count of 4.38. The geometric interpretation of content traits such as factualness, novelty, repetitiveness, and relevancy over a Hilbert vector space combines spatial volume (all trait importance) with vertices alignment (individual trait relevance) enabled the LLMaaJ to optimize on Human ToM. This resulted in an increase in content quality by extending the coverage of tennis action. Our work that was deployed at the US Open 2024 has been used across other live events within sports and entertainment.

**Concepts**

• Computing Methodologies • Artificial Intelligence • Natural language processing • Natural language generation

**Keywords**

Applied Computing, Generative AI, Sports and Entertainment, Agentic Frameworks, Theory of Mind

## 1 Introduction

The field of Generative Artificial Intelligence (GenAI) is rapidly enabling the creation of content generation workflows with a focus on agents. An AI agent is a program or system built with modern AI such as foundation models that can autonomously perform a task on behalf of another agent, program, or human [1, 2]. Individual LLMs that have increased with parameter size have shown very good capabilities on many applications but struggle on complex tasks [3, 4]. The performance over complex tasks can improve as the number of agents that work together increase [5]. Complex workflows, such as the US Open 2024 match reports, are built on top of agentic architectures where multiple agents work together on a common task [6, 7].

As workflows become more capable and prevalent in daily tasks, they must meet end user expectations in a trustworthy framework to be accepted within real-world scenarios. The social contract between AI agents and humans should reinforce familiarity and empathy [8]. The more familiar in knowledge up to the point of expert-level that a user has with an AI system, the more likely the acceptance rate [9]. However, the more familiar an AI system acts in behavior and meets the expectations of a user not bound by expert-level attainment, the more trust that can be earned [10]. Empathetic reasoning, content generation, and responses from LLMs are a major part of workflow effectiveness [11].

Through the Theory of Mind (ToM), which is a cognitive ability of humans to understand that other people have beliefs, intentions, and mental states, constructs can be designed to guide AI systems to align with user expectations [12]. This view can be thought of as Artificial Social Intelligence where multiple agents and humans work together on a cumulative task [13]. However, some work has shown that LLM's lack in the ability of understanding the intentions of others [14]. As work within this area continues, LLM's will begin to understand their users to create better human to generative AI interactions [15]. Social Intelligence methods and algorithms are being created to help the field pursue ToM [16].

Within our work, we explore the interaction of LLMs, LLMaaJ, and human content creators to optimize human mental alignment with generative AI. We began by investigating and evaluating how our techniques increased the usability of our generative AI system at the 2024 US Open for human content adjudicators [17, 18]. We used an agentic architecture where agents used the IBM Granite 13B chat LLM to generate factual bullet points about tennis matches [19]. Next, a Llama 3 70B LLM worked with the factual bullets to write fluent sections within a match report that summarized the most statistically significant situations about play. For each of the sections within the match report, a LLMaaJ in the form of another Llama 3 70B LLM evaluated the text based on several ToM dimensions such as novelty, repetitiveness, factualness, and relevance to the provided context and previous instruction within each of the prompts. An optimization function minimizes a geometric spatial loss that reduces the Hilbert space differences between a human's expected ToM dimensions with the results of the agentic workflow. During each iteration, a ToM Chain of Thought (CoT) along with a description of the spatial judgement is included within the next formed prompt to change the behavior of the agentic workflow [20]. This paper depicts our research and application of iteratively modifying generative AI content at the US Open 2024 to:

- Align human expectations with generative AI systems to decrease human manual content editing.
- Production of complex and long-form texts through CoT reasoning that limits anomalous output.
- ToM meta prompting techniques to enable LLM's to self-learn human preferences.
- Create large scale agentic generative AI workflows that balance prescriptive few-shot learning with creative context provided zero shot learning.

## 2 Related Works

The field of Generative AI and Human Computing Interaction is rapidly becoming a focal point within the overall field of AI. Entire workshops at the ACM CHI conferences have been focused on human digital interaction with generative AI [21]. Generative AI's presence in digital environments necessitates a reevaluation of how empathy is elicited, understood, and applied in both physical and digital realms that is influencing the design of platforms and technologies [22]. These bodies of work explore how humans understand, interact with, and interpret content created by LLMs and Foundation Models (FMs). When humans interact, a high level of empathy generally facilitates a trustworthy and fluid engagement that is central to social interaction [23]. One of the key aspects of this convergence is the concept of anthropomorphism, which is the tendency to attribute human-like qualities to non-human entities [Bojan, Strachan]. This trait is amplified by generative AI's ability to engage in human-like conversation and exhibit social awareness [23, 24]. As a result, AI systems need robust methods to assess and measure empathy, particularly in the context of generative models [25]. However, most LLM's and generative AI systems lack basic human social awareness or cognitive expectations that align with the ToM [23, 26]. Though independently, many works have explored Theory of Mind with respect to AI and Meta-Prompting, the two fields have not been combined into a comprehensive ToM CoT Meta-Prompting approach [27].

### 2.1 Theory of Mind

The ToM enables humans to understand that other humans or agents have their own mental states such as beliefs, intentions, and representations [28]. These traits are integral to human social success and the ability to predict others' behavior and response [28]. There have been many studies that explore different facets of Theory of Mind (ToM) in AI. Some authors research methods for AI agents to have empathy with each other. For example, Rabinowitz et al introduces a ToM neural network called ToMnet that learns to model the behavior of other agents through meta-learning that can pass ToM tasks such as the Sally-Anne false belief task [29]. This type of reasoning is extended to LLMaaJ within chatbots [30]. Within Zheng et al.'s work, LLM's evaluate other LLM responses to open questions to measure the similarity between the output and user preferences [30].

Other researchers focus on LLM's developing a sense of understanding human ToM, which emphasize that AI systems must exhibit ToM understanding to build trust and transparency with their human counterparts [29]. For example, an LLM-based evaluation framework for summarization tasks uses role-player prompting to simulate different perspectives of human thought with limitations on nuanced interpretation [31]. Additional dynamic ToM is defined as spontaneous ToM that aligns with the idea of creating LLMs capable of interpreting other model's AI output through a human-like evaluator rather than pre-programmed responses [32]. Going further, Williams explores the role of ToM in supporting Artificial Social Intelligence. This work presents Artificial Theory of Mind (AToM), in that agents need to be able to recognize and understand human mental states to collaborate effectively with their users [33]. Work by Jamali show experimental results that LLMs have initial evidence of maintaining capacity to infer another's perspective based on neural embeddings [34].

### 2.2 Meta Prompting

Meta-prompting refers to the technique of using an LLM to automatically generate or refine prompts that are fed back to itself or another LLM to enhance performance on downstream tasks. Through task agnosticism, meta-prompting draws a relationship between tasks, prompts, and LLM outputs so that models do not need task specific instructions [35]. For complex tasks, meta prompting can be used as a technique to improve the accuracy and robustness of LLMs [36]. Suzgun extends the idea of meta prompting by prompt tuning a complex task into smaller subtasks that are routed to specialized expert models [36]. Whereas Jiang et al uses a method of MetaPrompter, a meta-learning strategy, to improve few-shot text classification [37]. Further, LLM-as-a-Meta-Judge (LLMaaMJ) evaluates a model's judgements with a self-rewarding mechanism to refine judgment skills [38]. This unsupervised approach allows a model to jointly learn how to judge and to follow instructions within a prompt [30]. Interestingly, Kojima et al. define meta prompting as zero-shot CoT prompting. This work draws inspiration from the CoT method by appending instructions such as "Let's think step by step" to the input query, which encourages the model to generate a sequence of reasoning steps before arriving at the final answer [38].

### 2.3 Chain of Thought

The original CoT prompting that enhances the reasoning abilities of LLMs by breaking down complex problems into a sequence of intermediate steps was introduced by Wei et al [41]. The step-by-step reasoning process is similar to human problem-solving, which has led to substantial performance improvements across arithmetic, common sense, and symbolic reasoning [42, 43]. Much of CoT's effectiveness is attributed to its ability to guide LLMs through a logical progression that addresses some of the limitations of traditional few-shot learning methods. Further, the chain of reasoning from CoT allows full transparency to humans and other AI agents. The success of CoT has inspired a wave of research exploring variations and applications called Chain-of-X (CoX) paradigms [44, 45].

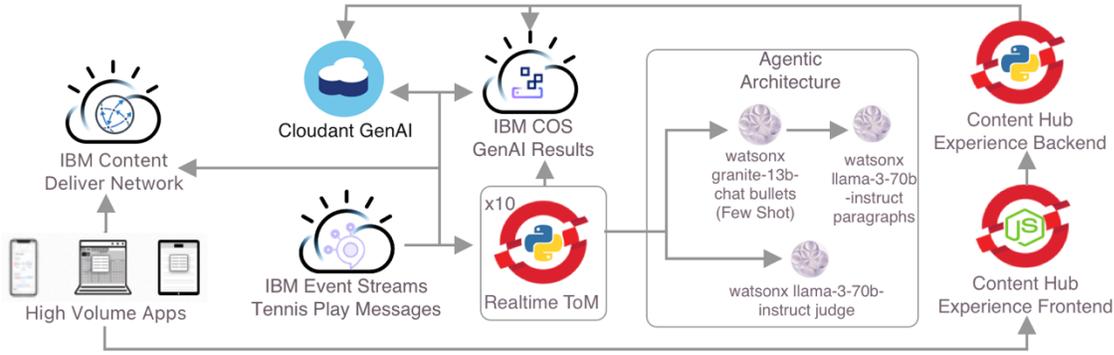

Figure 1: Realtime ToM Architecture deployed at the US Open 2024

However, despite the promising results from CoT, research suggests that the reasoning process in LLMs might not always reflect true causal understanding [46, 47, 48]. Several instances where LLMs generate correct answers despite incorrect reasoning steps indicate a potential reliance on spurious correlations within training data rather than genuine causal connections between steps [46, 47, 48]. Further, other works show that LLMs can generate diverse CoT paths for a single problem where each have a different degree of quality and effectiveness [49, 50]. Even so, leading industry models such as OpenAI's GPT-o utilize differing CoT paths to self-learn new skills called thinking [51].

## 2.4 Reinforcement Learning

When an LLM is deployed as an agent or within an operating environment, the model can learn through Reinforcement Learning techniques (RL), which is inspired by behavioral psychology [52]. In general, an agent performs an action within an environment to receive feedback in the form of a reward or penalty that updates a policy table. Traditional RL techniques are modeled as Markov Decision Processes (MDP) that run as Online RL, Offline RL, Imitation Learning (IL), Inverse RL, and Learning from Demonstrations (LfD) [53]. The combination of Deep Learning (DL) and RL has shown exceptional ability for agents to solve complex and high dimensional problems [54]. For example, one DL RL work mastered the game of Go while another beat professionals at Poker [55, 56].

Combining the techniques of RL with LLMs can create impressive results. DeepSeek-R1 and 6 dense models were distilled from Qwen and Llama [57]. These models showed SOTA performance on benchmark datasets such as AIME 2024, Codeforces, GPQA Diamond, MATH-500, MMLU, and SWE-bench Verified [57]. DeepSeek-R1-Zero evolved using RL from DeepSeekV3, which was further fine-tuned into DeepSeek-R1 [57]. Now, techniques such as RL with Human Feedback (RLHF) is a training phase with a human-in-the-loop to align the output of an LLM with the expectation of a user [53]. The RLHF can improve credit assignment, algorithmic design, and prompting [53]. Our paper addresses the problem of aligning LLM output with human expectations by optimizing prompts with RLHF.

## 3 Live Event and Data

Sports consumers ranging between casual and serious fans want to be engaged with the most interesting and latest information about their favorite athletes. The scale and speed of live events presents coverage and timeliness challenges for even the most seasoned production and broadcast teams. Our work focuses on increasing live event coverage at the US Open 2024 while maintaining high quality content that requires a small amount of human oversight. Over time, the amount of human content adjudication effort decreases as generative AI models align their output with human expectations.

With the speed of the tennis game encompassing over 255 matches that occur on 22 courts, sometimes at the same time, the deluge of data is overwhelming for fans, content producers, and computing systems. A prioritized list of 238 tennis statistics was tracked throughout a tennis match and summarized post-match by an agentic workflow depicted in Figure 2. Of the 254 matches, 239 of the post-match summaries utilized the ToM alignment optimization techniques. The anatomy of a post-match reports consists of an introduction describing the players, an action section showcasing tennis play, and a closing that recapped the match score. These sections were produced in parallel by 3 independent calls to the agentic workflow.

During play, scoring messages are sent to multi partition Kafka topics. As soon as a message is received by a consumer, a generative AI call is executed. For example, when a match completes, a message is sent to a topic's partition. A feature extractor and generative AI executor immediately processes the received message about the match to produce a post-match report. Over 803,000 unique users spent an average of 5.5 minutes reading the post-match reports.

## 4 Theory of Mind Formalization

During the creation of text, human editors work with the output of a generative AI system, $genAI(gllm, x_i)$, that encompasses a set, $j$, of large language models, $llm$, with input prompt, $x_i$.

$$gllm = \{\forall llm_j\} \qquad (1)$$

The initial output, $o_{i,k}$, from sample $i$ of the $llm$ pipeline is edited by a human editor, $h_k \in H$, that is drawn from a set of human editors, $H$. These human editors include subject matter experts for Tennis, IBM stakeholders, USTA stakeholders and professional content editors.

$$o_{i,k} = h_k\big(genAI(gllm, x_i)\big) \qquad (2)$$

Equation 2 depicts the results of generative text edits that will be judged by a generative pipeline, jllm. The vector $\vec{j_{1,k}}$ contains the set of orthogonal dimensions, $d_i$, that measures the confidence level that a property exists.

$$\vec{j_{1,k}} = genAI(jllm, o_{i,k}) \quad (3)$$

Now, each of the dimensions, $d_i$, is within the space $\mathbb{R}^{|d|}$ such that a square matrix, $e_k$ with shape $|d| \times |d|$, represents a human editor's k corrected expectations over all m points. To maintain orthogonality, a singular row, m, within $e_{i,m_k}$ lies on a singular dimension, n, such that all other dimensions are 0. However, to calculate covariance scaling and allow interdimensional relations, a raw matrix, $eraw_{i,m_k}$, does not reduce all other dimensions to 0 to maintain orthogonality.

$$e_{i,m_k} = \begin{bmatrix} d_{i,m} & \cdots & d_{i+y,m} \\ \cdot & d_{i+1,m+1} & \cdot \\ \cdot & & \cdot \\ d_{i,m+y} & \cdots & d_{i+y,m+y} \end{bmatrix}_k \quad (4)$$

$$\cos(d_m * \forall d_n) = 0 \therefore m \neq n \quad (5)$$

A scaled covariance matrix in equation 6, $scov_{i,m_k}$, between all combinations of dimensions, $d_i$, in $eraw_{i,m_k}$ determines the degree of relationships between dimensions.

$$scov_{i,m_k} = \begin{bmatrix} 1-|cov(d_{i,m}, d_{i,m})| & \cdot & 1-|cov(d_{i+y,m}, d_{i,m})| \\ \cdot & \cdot & \cdot \\ \cdot & \cdot & \cdot \\ 1-|cov(d_{i+y,m}, d_{i,m})| & \cdot & 1-|cov(d_{i+y,m+y}, d_{i+y,m+y})| \end{bmatrix}_k \quad (6)$$

The relationship between $e_{i,m_k}$ and $eraw_{i,m_k}$ will be through a lookup of the paired dimension $scov_{i,m_k}$ value. A mapping function will take the maximum $1 - |cov(d_{i,m}, d_{j,m})|$ across a column in $scov_{i,m_k}$. The value of the corresponding $e_{i,m_k}$ is the weighted edge between $(d_{i,m}, d_{j,m})$. Equation 7 shows a graph representation, $G(E, V)$, that takes the form of a polygon with edges E and vertices V.

$$G(E,V)_k; E = 1 - |cov(d_{i,m}, d_{j,m})|; V = e_{rows_k} \quad (7)$$

Now, a second set of generated output, $o_i$, without human correction that has been judged by an LLM pipeline, jllm, is prepared to create another polygon following equations 4, 6, 7.

$$o_i = genAI(llm, x_i) \quad (8)$$
$$\vec{j_i} = genAI(jllm, o_i) \quad (9)$$

Two graphs from the same input samples, $x_i$, can be compared with the ToM of area, $tma_k$, and the Theory of Mind distance, $tmd_k$. Equation 11 shows the differences between the areas of the polygons, $G$ and $G_k$, by taking the determinant of the $|d| \times |d|$ matrices created by the transformation function f. The $tmd_k$ is computed as the average of the cartesian distance between all coordinate nodes depicted in equation 12.

$$visualize(G(E,V)_k, G(E,V)) \quad (10)$$
$$tma_k = |det(f(G(E,V)_k)) - det(f(G(E,V)))| \quad (11)$$
$$tmd_k = Average\left(\sqrt{(G(E,V)_k - G(E,V))^2}\right) \forall V \quad (12)$$

In the case of rectangular matrices of the graphs $G$ and $G_k$, the Hausdorff measure can compute partial areas of the polygon. As shown in equation 13, for each dimension, $d_i$, is a Hausdorff

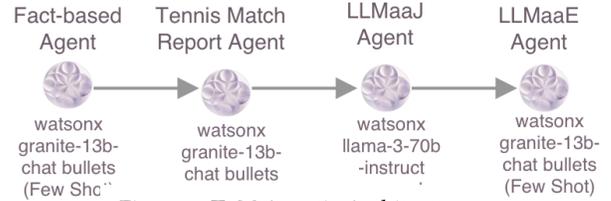

Figure 2: ToM Agentic Architecture

dimension that can be used to compute the size of each shape within a rectangular matrix.

$$Area_k \approx \frac{1}{2^{(|d|-1)}} * (volume_1 + \cdots + volume_d) \quad (13)$$

More generally, each matrix can be factored into multiple square matrices, $e_{i,m_k}'$ such that volumes of subspaces can be calculated.

$$volume_k = \iint \ldots \int |det(e_{i,m_k}')| dx_1 dx_2 dx_3 \ldots dx_k \quad (14)$$

After the $Area_k$ and original Area is known, equations 15 computes the difference in user's theory of mind, $tma_k$.

$$tma_k = |Area_k - Area| \quad (15)$$

Several parameters influence the output of an LLM. For example, each LLM prompt, $x_i$, contains an instruction, $i_z$, and an optional context, $c_z$. The instruction, $i_z$, tells the model a particular behavior to follow that establishes task following. Another parameter is called the temperature, t. Equation 16 shows the importance of temperature by scaling the probabilities assigned to each candidate word, $p_w$, where n is the size of the library and y is a candidate word. The smaller the temperature, the more confident a model is assigning probabilities to words.

$$p_w = \frac{e^{\frac{y_i}{t}}}{\sum_{k=1}^{n} e^{\frac{y_k}{t}}} \quad (16)$$

Another parameter is called the top-p, tp, which filters out words that fall below a cumulative distribution of the most probable words. Whereas the top-k, tk, considers only the kth highest probable words. Equation 17 shows our loss function while Equation 18 shows the definition of the Theory of the Mind, tom, as related to $tma_k$ and $tmd_k$. We now use both $tma_k$ and $tmd_k$ to calculate the overall judgement loss and define a threshold of optimality.

$$loss = \frac{\left(\frac{1}{2}*\left(\frac{tma_k}{Area_k}\right)^2 + \frac{1}{2}*abs\left(\frac{tma_k}{Area_k}\right)\right)}{2} + tmd_k \quad (17)$$

This loss is an equal combination of average of mean squared percentage error and absolute percentage error between the expected area and the current area and $tmd_k$

$$tom = \begin{cases} true: loss < 0.05 \\ false: otherwise \end{cases} \quad (18)$$

The optimization problem of Theory of the Mind is established by equation 19. Stochastic gradient descent steps from equation 20 and 21 with an instruction, $i_z$, held constant will continue until convergence. Next, another instruction, $i_z$, will go through the same process until the best combination of $i_z, tp,$ and $tk$ satisfy a maximization of equation 18. The empirical risk, $Q$, is an estimation of the probability of obtaining a $tma_{k_t}$ or $tmd_{k_t}$ value at step, t.

$$P_{tom}(tom = true|i_z, tp, tk) =$$
$$(tom = true|tma_k(i_z, tp, tk), tmd_k(i_z, tp, tk)) \quad (19)$$
$$tma_{k_{t+1}} = tma_{k_t} - \eta \Delta Q(tma_{k_t}) \quad (20)$$
$$tmd_{k_{t+1}} = tmd_{k_t} - \eta \Delta Q(tmd_{k_t}) \quad (21)$$

## 5 Theory of Mind Application

The personification of a content reviewer's mind aligns their expectations with the output of our generative AI workflow shown in Figure 1. The large-scale agentic architecture creates short and long form text for live events such as the US Open 2024. As tennis matches are scheduled and end, Kafka messages are sent to the Event Streams hub for the generation of either pre-match or post-match content. The Realtime ToM application is scaled out into 10 parallel processes. The IBM Granite 13B chat agent uses the few-shot learning pattern to finely control the types of bullet points that describe the tennis scene. The fact-based sentences are then transformed by a Llama 3 70B model into fluent paragraphs by a generic tennis writer persona agent. Next, another Llama 3 70B judges the paragraphs by the dimensions established by equation 4.

Through this process, each of the generative content pieces are stored within Cloudant, a document database, and Cloud Object Storage (COS), an object database. The content is loaded by a backend and rendered on a Content Hub experience. Content adjudicators log into the experience to edit or regenerate the tennis paragraphs. As content is edited, a user's expectation profile derived from equations 2, 3, and 4 is saved into Cloudant. Now the Realtime ToM can alter the instruction and context within the prompt to maximize the alignment of a human and LLM. Each iteration of the content generation workflow will use the modified instruction to streamline the publication of generative content with minimal human oversight. The published content is stored within a COS origin and fronted by a Content Delivery Network (CDN). The 14 million unique US Open users access the content through CDN caching layers. As a result, millions of users around the world can read articles created by a ToM agentic architecture workflow depicted in Figure 2.

### 5.1 Dimensions of Human Judgement

Throughout the US Open 2024 generative content production, several different types of errors impacted the quality of LLM output. The first one was inaccurate information around statistics. Within some inference calls, the models mixed together historical statistics such as first shot win percentage with real-time results. In other cases, numbers were fabricated or juxtaposed with the incorrect statistic. Another error type was model hallucination. In this case, a model would create untraceable assertions from source data about a tennis player or match. These types of errors required a lot of effort by human adjudicators because the novel text flowed very well with the rest of the correct text.

Within other instances, the quality of the text was very low. This happened when generative content was very repetitive about the same few statistics in distinct paragraphs. On the other extreme, a model could be rambling about too many unique statistics or even change topics to a subject not about tennis. Within these cases, content editors had to refocus the match reports through their own manual extractive summarization techniques. To add to the complexity, each evaluator had their own definition and measurements to determine generative text quality. As a result, personalized content edits were created by each individual reviewer.

To capture the human dimensions of human judgement, we worked with 4 content adjudicators to define 4 dimensions of evaluation:

1. Factualness = the content should be factually correct with the most recent information such as statistics.
2. Novelty = the content has creativity and adds extra context to the output.
3. Repetitiveness = the generated content discusses similar points in repetition.
4. Topic Alignment = each element within the text should have a direct relationship to a tennis match or player.

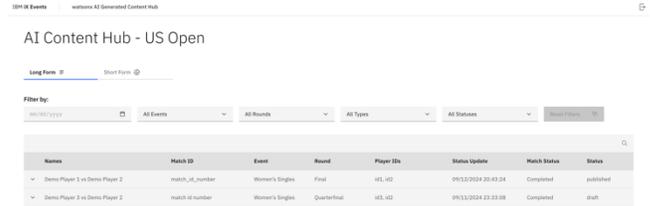

Figure 3: AI Content Hub table of tennis match content.

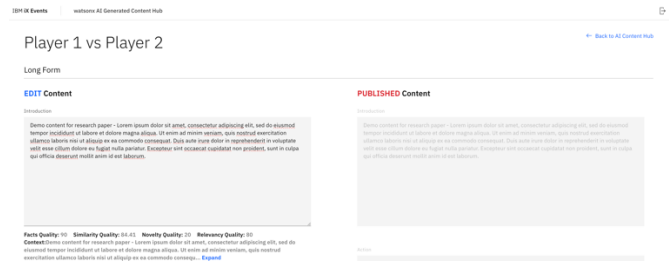

Figure 4: Generative AI content editing with the initial LLMaaJ

### 5.2 LLMs as a Judge

The identified qualitative dimensions such as Factualness, Novelty, Repetitiveness, and Topic Alignment are automatically extracted from tennis match reports by an LLM to align with corresponding dimensions of manual human judgement. The automatic judgement agent encapsulated a Llama 3 70B model combined with few-shot learning discerns a relation between a factual-based agent with the IBM Granite 13B Chat V2 models and a tennis match report writer agent with the IBM Granite 13B chat V2 model. The set of generated bullets by the fact-based agent becomes context input into the match report writer agent. Now, the LLM as a Judge agent measures each dimension of evaluation on a numerical score between 0 and 100. Table 1 provides the optimal scores for each dimension. However, each human editor has their own slightly different ideal dimensional scores. As part of the automatic LLMaaJ agent, the provided few shot examples define a response JSON format that can be consistently translated and post-processed by other parts of the workflow.

| Dimension | Ideal Score | Theoretical Ideal Alignment | Description |
|---|---|---|---|
| Factualness% | 100 | 0 | Factually accurate to the context provided, deducting points for each factual error. |
| Novelty% | 50 | 0 | A balance in creativity and expectations. |
| Repetitiveness% | 0 | 0 | Every statistic is unique and covered exactly once within the tennis match report. |
| Topic Alignment% | 100 | 0 | Highly relevant to the player and their tennis game. |

Table 1: Ideal dimensional scores for tennis match reports without personalization.

During the US Open 2024, the LLMaaJ agent analyzed 254 match reports twice. In the first pass, the LLMaaJ agent assesses each raw tennis match report on each of the four dimensions to create a 1024 score vector. In the second pass, the LLMaaJ agent analyzes human-edited tennis match reports to create a target 1024 score vector. The differences between $tma_k$ and $tmd_k$ provide a basis for the creation of a generated meta prompt. The 1024 score vector of human edited tennis match report is also used to create the covariance matrix.

### 5.3 LLMs as an Editor

The generation of a meta prompt is created by an LLM as an Editor (LLMaaE). The quantitative scores for each of the qualitative dimensions of both human edited and raw tennis match reports provided a foundation for prompt rewriting. The overall optimization problem depicted in Equations 17, 19, and 20 defines the halting criteria for acceptable tennis match reports for a given human reviewer's mental expectations. The explanation around the $tma_k$ and $tmd_k$ scores such as the most significant percentage differences between dimensions provided instructions for the reformulation of tennis match reports. For example, within the prompt rewriting task, the IBM Granite 13B Chat V2 model is provided with a series of descriptions within the general form of:

<dimension> is <delta>% <above/below> expectations. Please improve by <increasing/decreasing> <dimension>

Specific instructions for the LLMaaE define how much to change the article and in which direction.

1. "Novelty" has perfect expectation score. Do not change "novelty"
2. "Repetitive" is 10% above expectations. Please improve by decreasing repetitiveness.
3. "Factualness" is 20% below expectations. Please improve by increasing factualness.

The definition of each dimension is provided to the LLMaaE to provide clarity of how the LLMaaJ analyzed the tennis match report. These system prompts and instructions improve $tma_k$ and $tmd_k$ over time by enabling the LLMaaE to edit the tennis match reports to align towards human expectations. An example system prompt is shown below.

"*You are an Editor who re-writes the given paragraph based on the feedback to get it approved. Using the context and previously generated paragraph, write a new paragraph to improve the scores to meet the quality requirements. \n\n\n the parameters are based out of 0-100, \n\n\n Parameters: \n\n 1. Factualness - Full score if all the facts like numbers, names, gender pronouns etc. in paragraph are matching the context. \n 2. Novelty - Higher score if the individual sentence structure is changed, order is changed or new content is added. \n 3. Topic Alignment - High score if the paragraph tells the story of the context without missing any important facts. \n 4. Repetitiveness - Higher score if the paragraph has repetitive stats or talks about the same point again and again.*"

### 5.4 RLHF

Before a live event, such as the US Open tennis, the initial training of the reinforcement algorithm is performed by the LLMaaJ, which interprets the unedited tennis match reports. Over time, human editors modify the generative content to provide their content preferences. During this process, each human editor is assigned a profile model that contains their expected average novelty, repetitiveness, and factualness scores as described in section 5.3. To facilitate human feedback, an editor uses the AI Content Hub shown in Figure 1 to navigate through different generative AI content. After the editor selects a tennis match, a detailed view of the generative content shows the initial LLMaaJ dimensions of measurement. As depicted in Figures 3 and 4, the human editor can modify the content to change the novelty, repetitiveness, and factualness scores. Any number of dimensions of measurement can be added to support any content alignment type to an editor such as style, persona matching, and text complexity.

As an editor changes the content, the LLMaaJ produces new dimension values of evaluation that begin to align with the user's expectations. The average of each dimension is saved into a user's profile model. If additional facts are available and surfaced within a context field, a context similarity score shows the similarity between the generative AI text and ancillary facts. This feature enables users to pick fact-based content that they prefer, which could differ from an initial LLM.

The editor feedback is used by the LLMaaE to generate meta prompts. The optimization problem defined in section 4 aligns the dimensions of evaluation within a geometric and area space to a human editor's content expectations. Next, an LLM will generate content with the meta prompt, which is again evaluated by the LLMaaJ. The iterations toward aligning the LLM output with human expectations continue with the LLMaaE interpreting the LLMaaJ response by editing the prompt. The process continues until the alignment problem converges to a solution or processing time surpasses a time threshold determined by the human editor. This RLHF process is a powerful method that increases the relative content quality for editors.

### 6 Results

At the US Open 2024, human editors published articles and match summary content based on agentic workflows supported by IBM Granite and Meta Llama models shown in Figure 1. During their editorial process, the system learned editor mental models such that the content could be judged by LLMaaJ and edited by LLMaaE to align with human expectations. Table 1 shows the initial starting points of alignment within a 3-dimensional and 4-dimensional vector space. The percentage difference between the factual dimension perfectly matched between human and LLMs. However, the novelty percentage difference was very high at 66.6%. Human editors generally wanted more novelty and creativity added into the text by LLMs. Further, the relevance difference of -5.5% supports that human editors want the LLM pipeline to produce content that is a bit more relevant to the context of a tennis match. Within the 4-dimensional experiments, the average human editor asserted that the text was not repetitive enough.

Within a few experiments, the editors were presented with Natural Language Generation (NLG) templates of factual content. The extra information did not have a significant impact on the percentage differences across dimensions of alignment. Editors typically added a few facts to the initial generative AI content to increase the length of the articles, which maintained the relative measures of novelty, relevance, factualness, and repetitiveness.

| Initial Alignment | 3D with NLG | 3D without NLG | 4D with NLG | 4D without NLG |
|---|---|---|---|---|
| Initial Δfact% | 0.0 | 0.0 | 0.0 | 0.0 |
| Initial Δnovelty% | 66.6 | 66.6 | 66.0 | 66.6 |
| Initial Δrelevance% | -5.5 | -5.5 | -5.0 | -5.5 |
| Initial Δrepetitive% | NA | NA | -12.5 | -12.5 |

Table 1: The initial human ToM alignment with generative AI content without RLHF.

Human editors agreed on a maximum amount of time they would wait for the LLM pipeline to align to their expectations, which was 2 minutes. The 2-minute threshold mapped to 21 iterations on a shared watsonx Runtime environment. When the agentic framework did not converge, in the 3 dimensions of evaluation the factual dimension was worse than the initial starting point. However, novelty and relevance improved. Within the 4D dimensions, factualness remained constant between the initial starting point while relevance declined. Even so, repetitiveness and novelty improved. Each of the editors preferred to work with the non-converged rather than the initial generative AI content, despite the tradeoffs.

| Non-Convergence | 3D with NLG | 3D without NLG | 4D with NLG | 4D without NLG |
|---|---|---|---|---|
| Δfact% | 15.1 | 12.6 | -0.87 | -6.43 |
| Δnovelty% | 34.4 | 42.8 | 6.73 | 1.71 |
| Δrelevance% | -0.73 | 1.80 | 27.7 | 38.9 |
| Δrepetitive% | NA | NA | -2.60 | -2.91 |

Table 2: Human ToM alignment with a maximal LLMaaJ and LLMaaE non-converged iterations.

The tradeoffs are represented within Figures 5 and 6. The closer the shapes are to the origin, the better the optimization problem outlined in Section 4 converged. In another view, the smaller the spatial volume and coordinate matching between an editor's ToM model and the generative AI content's dimensions of measurement, the closer to alignment. Table 3 shows the results of ToM convergence.

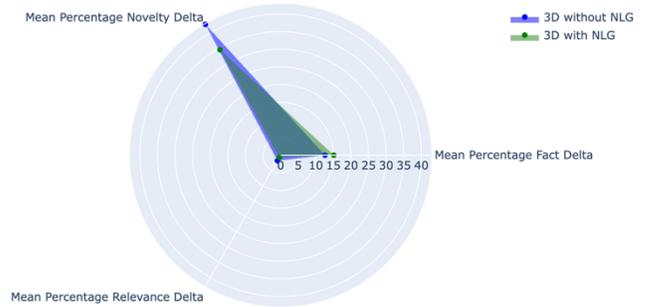

Figure 5: Non converged representation graph of 3D ToM

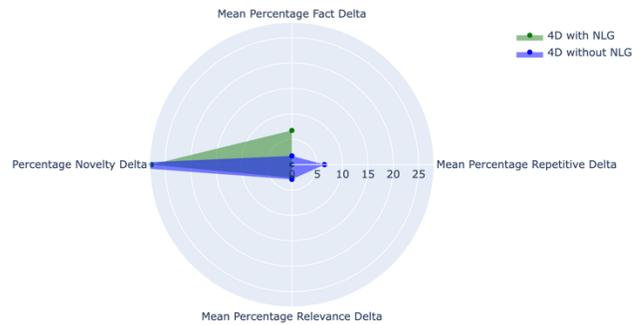

Figure 6: Non converged representation graph of 4D ToM

When the ToM alignment was solved and perfectly converged on 53.8% of cases, editors accepted and published the content with minimal or no text change. When human editors saw convergence, they were more likely to immediately publish generative AI content. The independence of each editor's ToM model increased the variety and variance between articles from differing editors.

| Convergence | 3D with NLG | 3D without NLG | 4D with NLG | 4D without NLG |
|---|---|---|---|---|
| Δfact% | 0.00 | 0.00 | 0.00 | 0.00 |
| Δnovelty% | 0.00 | 0.00 | 0.00 | 0.00 |
| Δrelevance% | 0.00 | 0.00 | 0.00 | 0.00 |
| Δrepetitive% | NA | NA | 0.00 | 0.00 |
| Convergence % | 53.0 | 58.3 | 43.1 | 60.9 |
| Average Convergence Iteration Number | 4.15 | 3.95 | 4.71 | 4.71 |
| Number of Samples | 239 | 239 | 239 | 239 |

Table 3: Human ToM alignment with converged LLMaaJ and LLMaaE RLHF.

## 7 Future Work

Some human editors wanted additional and specific dimensions of measurement. We would like to enable an editor to define their own dimensions, definitions, and measurement criteria to align towards their priorities. This type of method will jointly align both an editor's belief and ToM systems. Further, we would like to add a thinking mode, utilizing the thinking models, where editors can specify the maximum time the optimization algorithm will attempt a convergence. We would like to extend our work to follow the Condorcet Jury Theorem, which will add LLM jury members to the judge chamber. The majority rule is likely to be correct as long as each individual LLM Jury Member has a probability greater than 50% of making the correct decision. Finally, we would like to study the implications of sharing and mixing ToM profiles to create emergent styles around each dimension of thought.

## Acknowledgements

We would like to thank IBM Corporate Marketing and the MIT-IBM Watson AI Lab for their support during our investigations. Gratitude goes to our property partner The USTA (US Open) for enabling us to deliver live generative AI systems within their digital ecosystems.